\documentclass[runningheads]{llncs}
\usepackage{eccv}
\usepackage{eccvabbrv}
\usepackage{graphicx}
\usepackage{booktabs}
\usepackage[accsupp]{axessibility}  
\usepackage{hyperref}
\usepackage{orcidlink}
\usepackage{amsmath,amssymb}
\usepackage{multirow}
\usepackage[table]{xcolor}
\usepackage{algorithm}
\usepackage{algorithmic}
\usepackage{subcaption}
\usepackage{enumitem}
\usepackage{pifont}
\usepackage[most]{tcolorbox}
\newcommand{\ours}{MMAgent-R$^2$}
\definecolor{mygray}{gray}{.9}
\definecolor{brightorange}{HTML}{CC3300}
\definecolor{brightblue}{HTML}{0066CC}
\definecolor{brightgreen}{HTML}{336633}

\begin{document}

\title{\ours{}: Learning to Rerank and Reject for Agentic mRAG}

\author{
Tao Zhang$^*$\inst{1,2,3,4}\orcidlink{0009-0007-2963-2844} \and
Ziqi Zhang$^*$\inst{1,3}\orcidlink{0000-0002-5937-183X} \and
Zongyang Ma\inst{1,3}\orcidlink{0009-0003-1980-9379} \and
Yuxin Yang\inst{1,2,3}\orcidlink{0009-0000-3562-2369} \and
Bing Li$^\dagger$\inst{1,3,5}\orcidlink{0000-0001-6114-1411} \and
Chunfeng Yuan\inst{1,3}\orcidlink{0000-0003-2219-4961} \and
Kang Rong\inst{4}\orcidlink{0009-0004-8451-1146} \and
Fengyun Rao\inst{4}\orcidlink{0000-0002-2868-2088} \and
Jing LYU\inst{4}\orcidlink{0009-0004-2021-0256} \and
Weiming Hu\inst{1,2,3,6}\orcidlink{0000-0001-9237-8825}
}

\authorrunning{T.~Zhang et al.}

\institute{
State Key Laboratory of Multimodal Artificial Intelligence Systems, CASIA\\
\email{\{zhangtao2023,mazongyang2020,yangyuxin2023\}@ia.ac.cn}\\
\email{\{ziqi.zhang,bli,cfyuan,wmhu\}@nlpr.ia.ac.cn} \and
School of Artificial Intelligence, University of Chinese Academy of Sciences \and
Beijing Key Laboratory of Super Intelligent Security of Multi-Modal Information \and
WeChat Vision, Tencent Inc.\\
\email{\{rickrong,fengyunrao,eckolv\}@tencent.com} \and
PeopleAI Inc. \and
School of Information Science and Technology, ShanghaiTech University\\[3pt]
$^*$Equal contribution \quad $^\dagger$Corresponding author
}

\maketitle

\begin{abstract}
Knowledge-based Visual Question Answering (KB-VQA) requires models to retrieve visual entities matching the query image from large-scale encyclopedic knowledge bases and answer related questions. Existing multimodal Retrieval Augmented Generation (mRAG) methods rely on global visual features to match candidate entities, yet when the knowledge base contains numerous visually similar entities, the retriever struggles to distinguish them, populating the candidate set with visually similar but factually mismatched distractors. Since subsequent processing steps such as noise filtering are also confined to this fixed candidate set, errors from failed retrieval inevitably propagate to the final answer. To address these challenges, we propose \ours{}, an agentic mRAG framework that integrates visual reranking and active rejection as its internal verification mechanism. Visual reranking directly compares query and candidate images, capturing discriminative details beyond textual descriptions to precisely identify the target entity among similar candidates; active rejection discards unreliable results and retrieves additional candidates when no confident match is found, moving beyond the fixed candidate pool. We design a composite reward function with step-level verification rewards and achieve joint optimization of external retrieval, internal verification, and answer generation via GRPO training.  Experiments on InfoSeek, E-VQA, and MMhops demonstrate that \ours{} achieves state-of-the-art performance, with particularly notable advantages in challenging retrieval scenarios and complex multi-image multi-hop reasoning tasks.

\keywords{Knowledge-based VQA \and Retrieval-Augmented Generation \and Visual Agent \and Reinforcement Learning}
\end{abstract}

\begin{figure}[t]
    \centering
    \includegraphics[width=\textwidth]{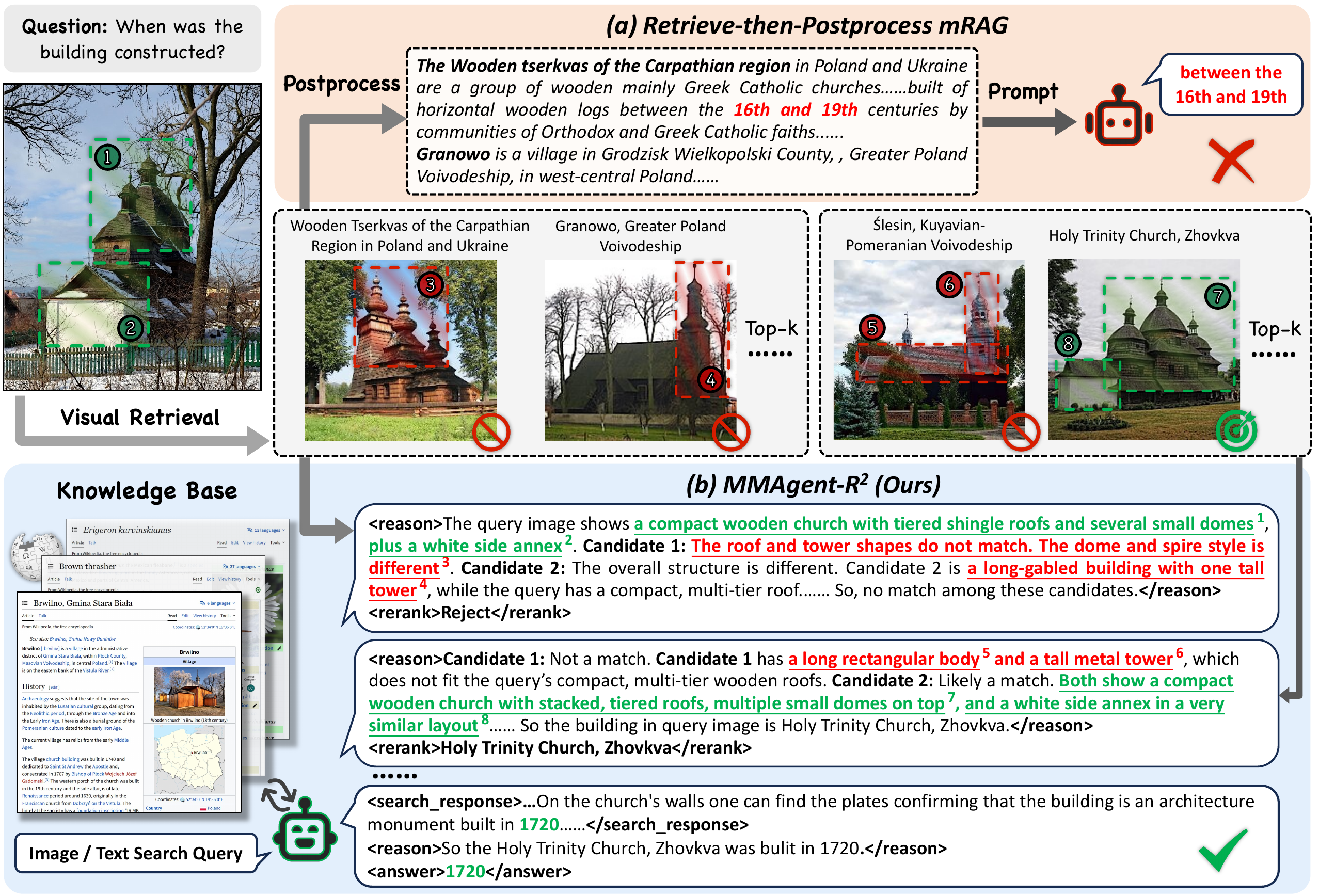}
    \caption{Comparison between the Retrieve-then-Postprocess mRAG and \ours{}. \textbf{(a)} With a fixed candidate set retrieved via global visual features, the model can be misled by visually similar yet mismatched entries. \textbf{(b)} \ours{} compares the query image against candidates (\ding{172}--\ding{179}) to rerank them, and rejects unreliable results to retrieve additional candidates when no confident match is found, enabling correct entity identification and answering.}
    \label{fig:teaser}
\end{figure}

\section{Introduction}
\label{sec:intro}

Knowledge-based Visual Question Answering (KB-VQA)~\cite{chen2023can, mensink2023encyclopedic, zhang2026mmhops} requires models to recognize visual entities in images and answer questions about their attributes. As illustrated in Fig.~\ref{fig:teaser}, answering ''When was this building constructed?'' hinges on precisely identifying the building and subsequently associating it with external knowledge (its construction year). Recently, multimodal Retrieval Augmented Generation (mRAG)~\cite{caffagni2024wiki,yan2024echosight,zhang2024mr,compagnoni2025reag,qi2024rora,hong2025knowledge,deng2025muka,ling2025mmkb,yuan2025mkg} methods have emerged as a promising approach for such tasks. These methods generally adopt a "Retrieve-then-Postprocess" paradigm, where the retrieval stage attempts to locate candidate entries visually similar to the query entity from a large-scale visual-textual encyclopedia knowledge base, and the post-processing stage then filters from a small, fixed set of candidates to identify content that accurately matches the query entity, thereby reasoning to generate the answer.

Despite their potential, existing mRAG methods encounter a critical identification bottleneck when distinguishing between visually similar entities. In the retrieval stage, existing methods primarily rely on global features from the full query image or cropped regions~\cite{hong2025knowledge, qiu2024snapntell} for similarity matching, which struggle to capture the key visual differences between similar entities (\eg, the tiered roof and dome configuration of the wooden churches in Fig.~\ref{fig:teaser}), populating candidates with visually similar yet factually mismatched entries. Although the post-processing stage reranks~\cite{yan2024echosight, ye2026qkvqa, compagnoni2025reag} textual paragraphs associated with candidate entities, textual descriptions inherently cannot represent subtle visual distinctions, rendering them ineffective at selecting the correct entity. Moreover, the reasoning scope is confined to a fixed candidate pool from the initial retrieval. If the correct entity is not recalled, the model is forced to reason over incorrect entities, inevitably propagating errors to answer generation. Meanwhile, such retrieve-then-postprocess pipelines typically depend on additional modules or predefined processing steps, hindering joint optimization and struggling with complex scenarios such as multi-image inputs and multi-hop reasoning.

To address these challenges, we propose \ours{}, an agentic mRAG framework integrating \textbf{visual reranking} and \textbf{active rejection}. These two capabilities constitute the model's internal verification mechanism, performing in-depth visual verification on each retrieved candidate batch. Visual reranking directly compares query and candidate images, capturing critical visual differences that textual descriptions fail to convey, thereby precisely identifying the target entity among similar candidates. When no reliable match exists, active rejection allows the model to discard the current results and retrieve additional candidates, moving beyond the fixed candidate pool and avoiding reasoning over incorrect entities. We unify reranking, rejection, and the invocation of image and text retrieval tools into a single reasoning model via reinforcement learning, eliminating dependence on additional modules. This agentic architecture naturally accommodates complex scenarios such as multi-image inputs and multi-hop reasoning, acquiring accurate knowledge through iterative retrieval and verification.

We train \ours{} via Group Relative Policy Optimization (GRPO)~\cite{shao2024deepseekmath} with a composite reward function: an outcome reward evaluates the correctness of the final answer, a format reward ensures compliance of structured outputs, and fine-grained verification rewards provide step-level supervision for each reranking and rejection decision, rewarding the model when it correctly identifies a matching candidate or correctly rejects when no match exists. This dense intermediate supervision enables the model to learn when to select and when to reject. Extensive experiments on InfoSeek~\cite{chen2023can}, E-VQA~\cite{mensink2023encyclopedic}, and MMhops~\cite{zhang2026mmhops} demonstrate that \ours{} achieves state-of-the-art performance across all benchmarks, with particularly significant gains on E-VQA (+7.2), which has the largest knowledge base and the highest retrieval difficulty, and on MMhops (Bridging +13.2, Comparison +10.4), which involves multi-image inputs and multi-hop reasoning, fully validating the effectiveness of our approach in challenging retrieval scenarios and complex reasoning tasks.

Our main contributions are summarized as follows:
\begin{itemize}[leftmargin=*]
    \item We propose \ours{}, an agentic mRAG framework in which \textbf{visual reranking} and \textbf{active rejection} jointly constitute the internal verification mechanism for external retrieval. The model autonomously searches, verifies, and answers within a multi-turn agent loop, without any additional modules.
    \item We design a composite reward function with step-level verification rewards that provide fine-grained supervision for each reranking and rejection decision, achieving joint optimization of external retrieval, internal verification, and answer generation via GRPO training.
    \item Extensive experiments show that \ours{} achieves state-of-the-art performance on InfoSeek, E-VQA, and MMhops, with particularly significant improvements in challenging retrieval scenarios and complex reasoning tasks.
\end{itemize}

\section{Related Work}
\label{sec:related}

\subsection{Knowledge-based VQA}
Knowledge-based Visual Question Answering (KB-VQA) requires models to recognize visual entities in images and leverage external knowledge bases to answer related questions. Early benchmarks such as OK-VQA~\cite{marino2019ok} and A-OKVQA~\cite{schwenk2022okvqa} focused on commonsense reasoning. Subsequent research shifted towards entity-centric settings with increasingly larger knowledge bases and finer entity granularity~\cite{lerner2022viquae, chen2023can}. Encyclopedic-VQA~\cite{mensink2023encyclopedic} further introduced two-hop reasoning that requires chaining multiple retrieval steps, and MMhops~\cite{zhang2026mmhops} extended this to multi-image inputs with multi-hop reasoning chains.

Multimodal Retrieval-Augmented Generation (mRAG) has emerged as the dominant framework for these tasks. Early approaches converted visual inputs into textual representations for text-level retrieval~\cite{lin2022retrieval}. Subsequent works shifted to image-level retrieval via visual encoders, including cross-modal retrieval that fuses image-to-image and image-to-text similarity for entity matching~\cite{lerner2024cross}, hierarchical retrieval that first recalls candidates through image similarity and then locates textual evidence~\cite{caffagni2024wiki}, cropped-region retrieval that reduces background interference to improve recall accuracy~\cite{hong2025knowledge, qiu2024snapntell}, and structured retrieval via multimodal knowledge graphs that organize unstructured knowledge into graph representations for more precise retrieval~\cite{yuan2025mkg}. Despite these advances, visual retrievers primarily rely on global feature matching, remaining prone to recalling visually similar yet factually mismatched candidates.

\subsection{Post-processing in mRAG}
After initial retrieval, how to effectively post-process candidates to reduce noise remains a key challenge for mRAG. Existing methods primarily operate at the textual evidence level. Representative approaches include leveraging VLMs to compute multimodal query-passage similarity for paragraph-level reranking~\cite{yan2024echosight}, employing reflection mechanisms that allow the model to autonomously assess paragraph relevance and decide whether to trigger supplementary retrieval~\cite{zhang2024mr}, question-guided filtering that injects question semantics into candidate encoding for cross-article chunk-level selection~\cite{ye2026qkvqa}, and multi-stage filtering pipelines that combine multimodal LLMs for hierarchical cross-filtering~\cite{hong2025knowledge} or train independent critic models to eliminate irrelevant texts~\cite{compagnoni2025reag}. Some works also leverage the model's own knowledge boundaries to dynamically generate semantic tags, filtering retrieved content based on its consistency with the model's existing knowledge~\cite{ling2025mmkb}. Although these methods have significantly advanced textual denoising, their filtering signals remain confined to the textual semantic space, struggling to distinguish fine-grained visual differences between highly similar entities. In contrast, \ours{} introduces visual reranking and active rejection for candidate images within a unified agentic framework, elevating post-processing from the textual to the visual level.

\section{Methodology}
\label{sec:method}

\begin{figure*}[t]
\centering
\includegraphics[width=\textwidth]{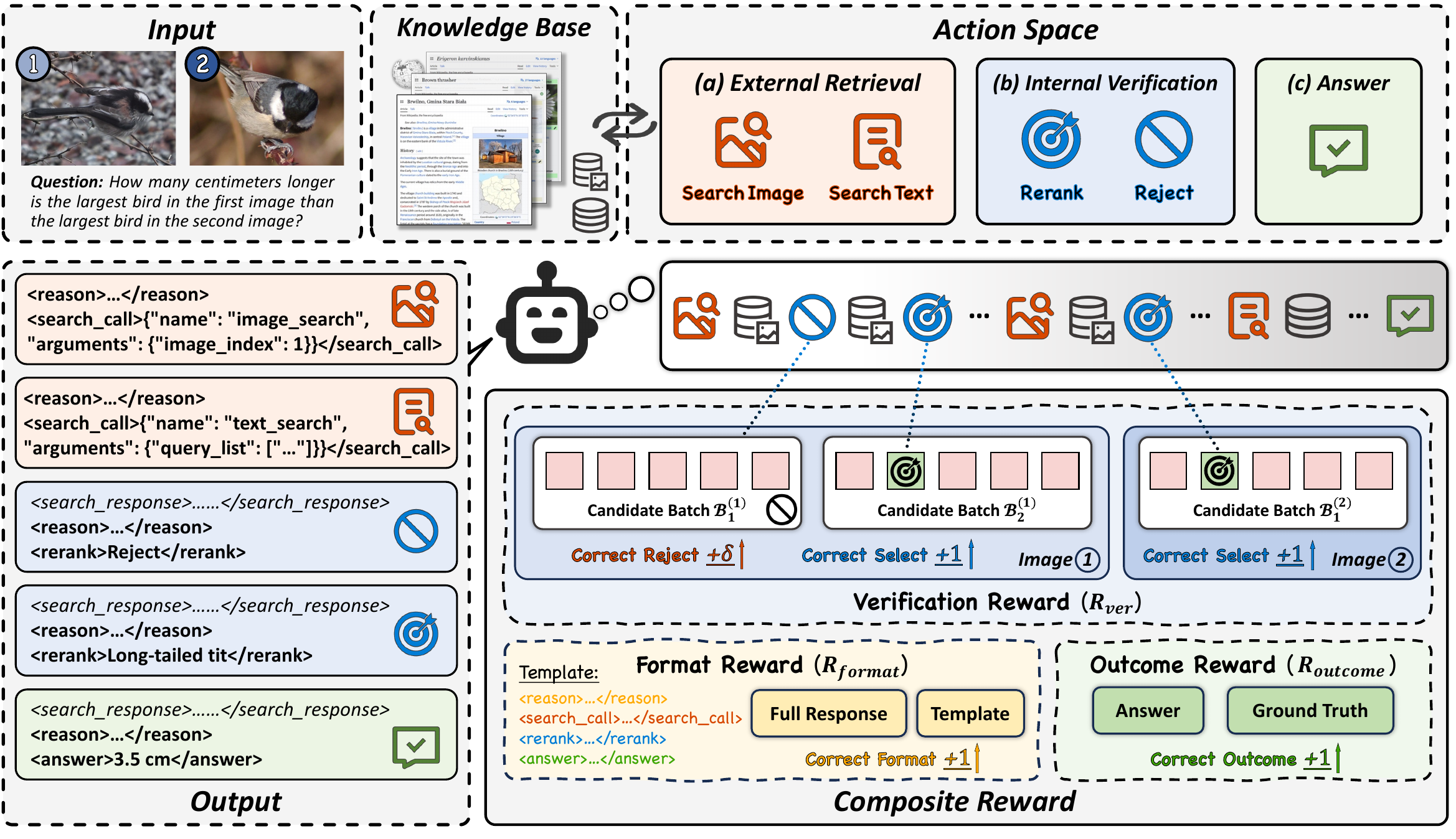}
\caption{\textbf{The overall framework of \ours{}.} Given multi-image inputs and a question, the VLM agent generates an interleaved ``Reason-then-Act'' trajectory by selecting and executing an operation at each step from an action space comprising \textcolor{brightorange}{External Retrieval}, \textcolor{brightblue}{Internal Verification}, and \textcolor{brightgreen}{Answer Generation}. The framework is optimized via RL with a composite reward function, where the verification reward provides dense step-level supervision for each reranking and rejection decision.}
\label{fig:framework}
\end{figure*}

\subsection{Problem Formulation}
\label{sec:formulation}
For the Knowledge-based Visual Question Answering (KB-VQA) task, given $N$ query images $\{I_q^1, \dots, I_q^N\}$ ($N \geq 1$) and a question $Q$, the model has access to an external knowledge base $\mathcal{KB}$ to extract relevant knowledge and generate an accurate answer $y$. Specifically, the knowledge base $\mathcal{KB}$ consists of a massive collection of structured web pages (\eg, Wikipedia). We formalize each entry as a triplet $(E, \mathcal{I}_e, \mathcal{T}_e)$, where $E$ represents an entity, $\mathcal{I}_e$ is the set of reference images associated with the entity, and $\mathcal{T}_e$ is the set of detailed textual paragraphs describing the entity. In practice, the images and texts across all entries in the knowledge base constitute the visual knowledge base $\mathcal{KB}_v = \bigcup_e \mathcal{I}_e$ and the textual knowledge base $\mathcal{KB}_t = \bigcup_e \mathcal{T}_e$ used for retrieval.

\subsection{\ours{}}
As illustrated in Fig.~\ref{fig:framework}, we formulate the reasoning process as a multi-turn agentic decision-making process. We define the VLM as an agent driven by a policy $\pi_\theta$ and construct an action space $\mathcal{A}$ comprising ``external retrieval'', ``internal verification'', and ``answer generation''. Specifically, the agent can invoke retrieval tools to recall candidate images from $\mathcal{KB}_v$ or extract relevant text from $\mathcal{KB}_t$. Concurrently, based on its perception and reasoning capabilities, it can perform precise "visual reranking" or "active rejection" on the recalled candidate images. Conditioned on the interaction history $\mathcal{H}_t$, the policy $\pi_\theta$ autonomously generates a sequence of thoughts and actions $a_t \in \mathcal{A}$ within a maximum of $T_{\max}$ turns. Through iterative exploration and verification, the agent ultimately executes the answer generation action to output the answer $y$.

\begin{table}[htbp]
\centering
\caption{Summary of the Agent Action Space $\mathcal{A}$. The policy $\pi_\theta$ can execute \textcolor{brightorange}{External Retrieval} actions to gather knowledge, perform \textcolor{brightblue}{Internal Verification} actions to validate retrieved candidates, or trigger \textcolor{brightgreen}{Answer Generation} to conclude the task.}
\label{tab:action_space}
\small
\resizebox{\textwidth}{!}{
\begin{tabular}{lll}
\toprule
\textbf{Category} & \textbf{Action} & \textbf{Model Output \& Environment Feedback} \\
\midrule
\multirow{2}{*}{\shortstack[l]{\textbf{External} \\ \textbf{Retrieval}}} 
& $\texttt{SearchImage}(j)$ & The model specifies query image index $j$. The environment returns a batch of Top-$K$ \\
& & visually similar candidates from $\mathcal{KB}_v$; consecutive calls yield non-overlapping batches. \\
\cmidrule{2-3}
& $\texttt{SearchText}(\{q_i\}_{i=1}^m)$ & The model generates a list of $m$ text queries. For each query $q_i$, the environment \\
& & retrieves and concatenates the Top-$K$ most relevant paragraphs from $\mathcal{KB}_t$. \\
\midrule
\multirow{2}{*}{\shortstack[l]{\textbf{Internal} \\ \textbf{Verification}}} 
& $\texttt{Rerank}(e_k)$ & The model outputs entity name $e_k$ from the current batch via visual comparison. \\
& & The environment records this entity for subsequent retrieval and reasoning. \\
\cmidrule{2-3}
& $\texttt{Reject}()$ & The model outputs the Reject signal, determining no match in the current batch. \\
& & The environment triggers $\texttt{SearchImage}$ to provide the next batch (up to $\text{Rej}_{\max}$ times). \\
\midrule
\multirow{2}{*}{\shortstack[l]{\textbf{Answer} \\ \textbf{Generation}}}
& $\texttt{Answer}(y)$ & The model synthesizes gathered evidence and generates the final answer $y$. \\
& & The environment terminates the reasoning loop. \\
\bottomrule
\end{tabular}
}
\end{table}

\subsubsection{3.2.1 Action Space}
\label{sec:tools}
As detailed in Table~\ref{tab:action_space}, we structure the agent's action space $\mathcal{A}$ into three categories: external retrieval for gathering multimodal knowledge, internal verification driven by the model's intrinsic reasoning, and answer generation for concluding the task.

\noindent \textbf{External Retrieval.} These actions enable the agent to query the external knowledge base to gather the multimodal evidence required for problem-solving.
\begin{itemize}[leftmargin=*]
    \item \textbf{Image Retrieval ($\texttt{SearchImage}(j)$):} This action encodes the query image $I_q^j$ and searches a pre-built feature index for similar candidate entries. Given the query index $j$, the environment returns a batch $\mathcal{B} = \{(I_k, e_k)\}_{k=1}^{K}$ comprising $K$ candidate entries, each containing an image and its entity name. Crucially, this action supports dynamic candidate expansion: consecutive calls for the same query image yield non-overlapping subsequent batches, enabling the model to break through the limitations of the initial Top-$K$ results and continuously explore the candidate space.
    \item \textbf{Text Retrieval ($\texttt{SearchText}(\{q_i\}_{i=1}^m)$):} The model leverages this action to extract relevant descriptive evidence from the textual knowledge base $\mathcal{KB}_t$. To enhance information acquisition efficiency, this action supports the parallel submission of multiple queries $\{q_1, \dots, q_m\}$. The environment retrieves paragraphs relevant to each query and returns the concatenated results. This mechanism allows the model to autonomously aggregate multi-dimensional textual facts within a single interaction turn, providing comprehensive evidentiary support for the final answer generation.
\end{itemize}

\noindent \textbf{Internal Verification.}
\ours{} leverages the fine-grained perception and reasoning capabilities of the VLM to conduct deep internal verification on the externally retrieved candidate batches, thereby precisely identifying the target entity among highly similar candidate images. Upon receiving the current candidate batch $\mathcal{B}_t^{(j)}$, the policy $\pi_\theta$ performs a direct visual comparison between the query image and all candidate images. Conditioned on the cumulative interaction history $\mathcal{H}_t$, the model generates a decision action:
\begin{equation}
a_t = \pi_\theta\!\left(I_q^j,\; \mathcal{B}_t^{(j)},\; \mathcal{H}_t\right) \in \left\{\texttt{Rerank}(e_1), \ldots, \texttt{Rerank}(e_K),\; \texttt{Reject}()\right\}
\end{equation}
\begin{itemize}[leftmargin=*]
    \item \textbf{Visual Reranking ($\texttt{Rerank}(e_k)$):} When the model verifies through visual feature comparison that entity $e_k$ in the candidate batch accurately matches the query image, it executes this reranking action and outputs $e_k$. This entity name is then appended to the context history, serving as a definitive premise for subsequent reasoning.
    \item \textbf{Active Rejection ($\texttt{Reject}()$):} When the model determines that no candidate image in current batch matches the query image, it executes this action. This action triggers dynamic candidate expansion, and the environment provides the next batch of candidate images. To balance search depth and reasoning efficiency, we set a maximum rejection count $\text{Rej}_{\max}$ for each query image.
\end{itemize}
By integrating ``reranking'' and ``rejection'' as the agent's internal verification actions, \ours{} achieves strict verification of retrieval results: ``reranking'' precisely identifies the target among similar retrieval results, while ``rejection'' dynamically expands the search boundary when initial retrievals fail. Working in synergy, they enable the model to autonomously explore complex retrieval spaces, ensuring that final reasoning and answer generation remain grounded in definitive visual evidence.

\noindent \textbf{Answer Generation.}
Based on the accumulated interaction history $\mathcal{H}_t$, the agent executes this terminal action to synthesize the gathered evidence and generate the final answer $y$, thereby concluding the reasoning process.

\subsubsection{3.2.2 Agentic Reasoning Loop}
\label{sec:agent_loop}
\ours{} operates through a multi-turn ``Reasoning-Action-Feedback'' paradigm. Each iteration of the loop involves the following logic:
\begin{itemize}[leftmargin=*]
    \item \textbf{Reasoning and Action:} The agent performs autonomous reasoning based on the current interaction history $\mathcal{H}_{t-1}$ and selects the next action. Instead of following a fixed sequence of steps, it flexibly decides, based on the reasoning progress, whether to initiate external searches to acquire new clues or to perform internal verification (\texttt{Rerank}/\texttt{Reject}) to examine the reliability of external information.
    \item \textbf{Environment Feedback:} Upon executing the selected action $a_t$, the environment returns a corresponding observation $o_t$ (such as image batches or textual facts). This result, along with the action command, is synchronously appended to the conversation history, forming a continuously updated multimodal evidence chain to support subsequent decisions.
    \item \textbf{Termination and Boundaries:} The loop terminates when the agent provides the final answer $\texttt{Answer}$ or reaches the maximum turn limit $T_{\max}$. Furthermore, the system monitors the exploration depth for each individual query image; once the maximum rejection limit $\text{Rej}_{\max}$ is reached, the environment provides a boundary prompt to guide the agent in achieving a balance between search scope and reasoning efficiency.
\end{itemize}
This framework overcomes the limitations of static pipelines and supports highly autonomous reasoning paths. The agent can flexibly schedule retrieval, reranking, and rejection actions based on the context, or even answer directly using its parametric knowledge. Through reinforcement learning, the model can learn to autonomously optimize interaction strategies within complex reasoning paths in an end-to-end manner.

\subsection{Agentic RL Training}
\label{sec:training}
We train \ours{} using Group Relative Policy Optimization (GRPO). For each training sample consisting of query images $\{I_q^j\}_{j=1}^N$, question $Q$, ground-truth answer $y^*$, and the corresponding ground-truth entities $\{e^{*,j}\}_{j=1}^N$, we sample $G$ complete multi-turn rollouts from the current policy $\pi_\theta$ and employ a carefully designed reward function to guide the model toward learning the optimal decision-making strategy.

\subsubsection{3.3.1 Composite Reward Function}
To supervise the agent's complex behaviors across multiple dimensions, we define a composite reward function:
\begin{equation}
\label{eq:reward}
R = R_{\text{outcome}} + R_{\text{format}} + R_{\text{ver}}
\end{equation}
\begin{itemize}[leftmargin=*]
    \item \textbf{Outcome Reward ($R_{\text{outcome}}$):} Evaluates the correctness of the final output $y$. It yields 1 if the answer matches the ground truth $y^*$, and 0 otherwise.
    \item \textbf{Format Reward ($R_{\text{format}}$):} Verifies whether the output of each turn strictly adheres to the predefined action syntax. This reward ensures stable structured interaction between the agent and the environment.
    \item \textbf{Verification Reward ($R_{\text{ver}}$):} Serving as the signal to guide the agent in learning internal verification behaviors, we provide step-level supervision for every verification decision ($\texttt{Rerank}$ or $\texttt{Reject}$) across all query images:
    \begin{equation}
    \label{eq:ver_reward}
    R_{\text{ver}} = \sum_{j=1}^{N}\; \sum_{t=1}^{T_j} \rho\!\left(a_t^{(j)},\; \mathcal{B}_t^{(j)},\; e^{*,j}\right)
    \end{equation}
    where $e^{*,j}$ is the ground-truth entity for the $j$-th image, $T_j$ denotes the number of verification steps, and $a_t^{(j)}$ is the action taken for the candidate batch $\mathcal{B}_t^{(j)}$. The per-step reward $\rho$ is defined as:
    \begin{equation}
    \label{eq:step_reward}
    \rho(a, \mathcal{B}, e^*) = 
    \begin{cases} 
    \delta & \text{if } a = \texttt{Reject} \text{ and } e^* \notin \mathcal{B} \\ 
    1 & \text{if } a = \texttt{Rerank}(e^*) \\ 
    0 & \text{otherwise}
    \end{cases}
    \end{equation}
    Here, $\delta$ represents the reward weight for correct rejections. This dense supervision mechanism not only complements the sparse final answer signal but also explicitly encourages the model to choose ``rejection'' over ``guessing'' when faced with uncertain candidate batches.
\end{itemize}

\subsubsection{3.3.2 Environment Response Masking}
During multi-turn rollout training, we mask all tokens corresponding to environment responses (\eg, retrieved images, textual paragraphs). This operation ensures that policy gradient updates are applied exclusively to the "Reasoning" and "Action" tokens generated by the model itself, ensuring the model remains strictly accountable for its own autonomous decision-making.

\section{Experiments}
\label{sec:experiments}

\subsection{Experimental Setup}
\label{sec:setup}

\subsubsection{4.1.1 Datasets and Evaluation Metrics}
We train and evaluate our model on three knowledge-based Visual Question Answering (VQA) benchmarks. 

\noindent \textbf{InfoSeek}~\cite{chen2023can} contains approximately 1.3M image-question-answer triplets, covering 11K Wikipedia entities and 2.7K entity categories, with answers classified into string, time, and numerical types. It defines two evaluation subsets: Unseen Question and Unseen Entity. For evaluation, string and time types use VQA accuracy (exact match after normalization, with a $\pm 1$ year tolerance for time); the numerical type uses Relaxed Accuracy (a prediction is correct if it falls within a 10\% tolerance range or achieves an IoU $\geq 50\%$ for range answers). The overall accuracy is the harmonic mean of these two subsets. Following previous setups~\cite{yan2024echosight, zhang2024mr}, we adopt a knowledge base comprising 100K Wikipedia pages.

\noindent \textbf{Encyclopedic-VQA (E-VQA)}~\cite{mensink2023encyclopedic} consists of 221K independent question-answer pairs, each equipped with up to 5 images, yielding approximately 1M VQA triplets. It covers 16.7K fine-grained categories sourced from iNaturalist~\cite{van2018inaturalist} and Google Landmarks~\cite{weyand2020google}, featuring single-hop and two-hop reasoning questions. Accuracy is measured using the BEM~\cite{bulian2022tomayto} (BERT Matching) score, where a prediction is correct if its score is $\geq 0.5$. We utilize the provided knowledge base containing 2M Wikipedia pages.

\noindent \textbf{MMhops}~\cite{zhang2026mmhops} is a benchmark specifically designed for multimodal multi-hop reasoning, containing 31.1K samples divided into Bridging (single-image chained reasoning) and Comparison (multi-image comparative reasoning) tasks. All samples require 3 to 4 reasoning steps to deeply integrate visual and textual knowledge. It adopts the same evaluation protocol as InfoSeek~\cite{chen2023can}. Following the benchmark's default setting, we employ a Wikipedia knowledge base of 100K pages.

\subsubsection{4.1.2 Implementation Details}
\label{sec:implementation}

We implement \ours{} on two base models: Qwen2.5-VL-7B-Instruct~\cite{bai2025qwen25vltechnicalreport} and Qwen3-VL-8B-Instruct~\cite{bai2025qwen3}. For image retrieval, we encode the images and titles of Wikipedia pages into combined features using EVA-CLIP-8B~\cite{sun2023eva} and build a candidate index with Faiss-GPU~\cite{douze2025faiss}. Given a query image, we encode it with the same model and retrieve the top-$K$ most similar candidates by cosine similarity, returning them as a single batch with a default size of $K{=}5$. Each query image is allowed up to $\text{Rej}_{\max}{=}2$ rejections by default, with each rejection triggering retrieval of the next candidate batch. For text retrieval, we build a vector index with E5~\cite{wang2022text} over Wikipedia articles segmented into 800-character chunks, returning the top-3 most relevant passages per query. We sample 65K, 67K, and 17K training instances from the training splits of InfoSeek, E-VQA, and MMhops respectively, rather than using the full datasets. Training is conducted with the GRPO algorithm~\cite{shao2024deepseekmath} on the verl framework~\cite{sheng2024hybridflow}, using a global batch size of 1024 for 2 epochs, with $G{=}8$ rollout trajectories per instance. The learning rate is set to $1 \times 10^{-6}$ and the maximum number of interaction turns is capped at $T_{\max}{=}8$. The default reward weight for correct rejection is $\delta{=}0.2$. Following DAPO~\cite{yu2025dapo}, we omit the KL divergence penalty to improve training efficiency.

\begin{table}[t]
\centering
\caption{VQA accuracy on E-VQA~\cite{mensink2023encyclopedic} and InfoSeek~\cite{chen2023can}. $^\dagger$Results obtained with a different knowledge base and thus not directly comparable.}
\label{tab:1}
\small
\resizebox{\textwidth}{!}{
\begin{tabular}{@{}lllccccc@{}}
\toprule
\multicolumn{1}{c}{\multirow{2}{*}{\textbf{Method}}} & \multicolumn{1}{c}{\multirow{2}{*}{\textbf{Model}}} & \multicolumn{1}{c}{\multirow{2}{*}{\textbf{Retriever}}} & \multicolumn{2}{c}{\textbf{E-VQA}} & \multicolumn{3}{c}{\textbf{InfoSeek}} \\
\cmidrule(lr){4-5} \cmidrule(lr){6-8}
 & & & Single-Hop & All & Unseen-Q & Unseen-E & All \\ 
\midrule
\multicolumn{8}{l}{\textit{\textbf{Direct Answer}}} \\
InstructBLIP~\cite{dai2023instructblip} & Flan-T5$_{\text{XL}}$ & - & 11.9 & 12.0 & 8.9 & 7.4 & 8.1 \\ 
BLIP-2~\cite{li2023blip} & Flan-T5$_{\text{XL}}$ & - & 12.6 & 12.4 & 12.7 & 12.3 & 12.5 \\ 
GPT-4V~\cite{achiam2023gpt} & - & - & 26.9 & 28.1 & 15.0 & 14.3 & 14.6 \\ 
Qwen2.5-VL-7B~\cite{bai2025qwen25vltechnicalreport} & Qwen2.5-VL-7B & - & 19.0 & 18.8 & 19.7 & 19.4 & 19.6 \\
Qwen3-VL-8B~\cite{bai2025qwen3} & Qwen3-VL-8B & - & 20.3 & 20.5 & 20.8 & 18.5 & 19.6 \\ 
\midrule
\multicolumn{8}{l}{\textit{\textbf{Multimodal RAG}}} \\
DPR$_{\text{v+t}}$$^\dagger$~\cite{lerner2024cross} & Multi-passage BERT & CLIP ViT-B/32 & 29.1 & - & - & - & 12.4 \\
RORA-VLM$^\dagger$~\cite{qi2024rora} & LLaVA-v1.5-7B & CLIP ViT-L/14 & - & 20.3 & 25.1 & 27.3 & - \\
CoMEM$^\dagger$~\cite{wu2025towards} & Qwen2.5-VL-7B & Custom VLM & - & - & 32.8 & 28.5 & - \\
Wiki-LLaVA~\cite{caffagni2024wiki} & LLaVA-v1.5-7B & CLIP ViT-L/14 & 17.7 & 20.3 & 30.1 & 27.8 & 28.9 \\ 
mKG-RAG~\cite{yuan2025mkg} & LLaVA-MORE-8B & Custom VLM & 38.4 & 36.3 & 41.4 & 39.6 & 40.5 \\ 
EchoSight~\cite{yan2024echosight} & Mistral-7B/LLaMA3-8B & EVA-CLIP-8B & 41.8 & - & - & - & 31.3 \\
mR$^2$AG~\cite{zhang2024mr} & LLaVA-v1.5-7B & CLIP ViT-L/14 & - & - & 40.6 & 39.8 & 40.2 \\
MMKB-RAG~\cite{ling2025mmkb} & Qwen2-7B & EVA-CLIP-8B & 39.7 & 35.9 & 36.4 & 36.3 & 36.4 \\ 
VLM-PRF~\cite{hong2025knowledge} & InternVL3-8B & EVA-CLIP-8B & 40.1 & 39.2 & 43.5 & 42.1 & 42.5 \\
QKVQA~\cite{ye2026qkvqa} & LLaMA-3.1-8B & EVA-CLIP-8B & 49.5 & - & 45.0 & 43.5 & 44.2 \\ 
QKVQA~\cite{ye2026qkvqa} & Qwen2.5-VL-7B & EVA-CLIP-8B & \underline{53.7} & - & 38.2 & 37.6 & 37.9 \\ 
ReAG~\cite{compagnoni2025reag} & Qwen2.5-VL-7B & EVA-CLIP-8B & 44.9 & \underline{47.0} & \underline{48.3} & \underline{46.2} & \underline{47.2} \\
\midrule
\multicolumn{8}{l}{\textit{\textbf{Agentic mRAG}}} \\
MMhops-R1~\cite{zhang2026mmhops} & Qwen2.5-VL-7B & CLIP ViT-L/14 & - & - & 33.8 & 32.6 & 33.2 \\
\rowcolor{mygray}
\textbf{\ours{}} & Qwen2.5-VL-7B & EVA-CLIP-8B & 55.4{\scriptsize\color{blue}{\,$\uparrow$1.7}} & 53.6{\scriptsize\color{blue}{\,$\uparrow$6.6}} & \textbf{50.7}{\scriptsize\color{blue}{\,$\uparrow$2.4}} & \textbf{49.7}{\scriptsize\color{blue}{\,$\uparrow$3.5}} & \textbf{50.2}{\scriptsize\color{blue}{\,$\uparrow$3.0}} \\
\rowcolor{mygray}
\textbf{\ours{}} & Qwen3-VL-8B & EVA-CLIP-8B & \textbf{55.9}{\scriptsize\color{blue}{\,$\uparrow$2.2}} & \textbf{54.2}{\scriptsize\color{blue}{\,$\uparrow$7.2}} & 49.0{\scriptsize\color{blue}{\,$\uparrow$0.7}} & 49.2{\scriptsize\color{blue}{\,$\uparrow$3.0}} & 49.1{\scriptsize\color{blue}{\,$\uparrow$1.9}} \\
\bottomrule
\end{tabular}
}
\end{table}

\begin{table}[t]
\centering
\setlength{\tabcolsep}{3pt}        
\renewcommand{\arraystretch}{1.0} 
\begin{minipage}[t]{0.49\linewidth}
\centering
\caption{VQA accuracy on MMhops~\cite{zhang2026mmhops}. \ours{}-7B and \ours{}-8B are initialized from Qwen2.5-VL-7B and Qwen3-VL-8B, respectively.}
\label{tab:mmhops}
\scriptsize
\begin{tabular}{lcc}
\toprule
Method & Bridging & Comparison \\
\midrule
Gemini-2.5-flash~\cite{comanici2025gemini}  & 46.6 & 23.2 \\
Gemini-2.5-pro~\cite{comanici2025gemini}  & \underline{54.0} & \underline{29.4} \\
EchoSight~\cite{yan2024echosight} & 13.6 & 4.81 \\
OmniSearch~\cite{li2024benchmarking} & 42.7 & 17.0 \\
MMhops-R1~\cite{zhang2026mmhops} & 51.4 & 22.0 \\
\rowcolor{mygray}
\textbf{\ours{}-7B} & 65.9{\scriptsize\color{blue}{\,$\uparrow$11.9}} & 34.9{\scriptsize\color{blue}{\,$\uparrow$5.5}} \\
\rowcolor{mygray}
\textbf{\ours{}-8B}  & \textbf{67.2}{\scriptsize\color{blue}{\,$\uparrow$13.2}} & \textbf{39.8}{\scriptsize\color{blue}{\,$\uparrow$10.4}} \\
\bottomrule
\end{tabular}
\end{minipage}\hfill
\begin{minipage}[t]{0.49\linewidth}
\centering
\caption{Entity identification accuracy (Ident.\%) of \ours{} against retriever recall. R@15 represents candidate pool coverage, \ie, the upper bound achievable by reranking and rejection.}
\label{tab:retrieval}
\scriptsize
\setlength{\tabcolsep}{1pt}
\begin{tabular}{lcccc}
\toprule
 & \multirow{2}{*}{InfoSeek} & \multirow{2}{*}{E-VQA} & \multicolumn{2}{c}{MMhops} \\
\cmidrule(lr){4-5}
 & & & Bri. & Comp. \\
\midrule
R@1 & 53.7 & 15.4 & 50.1 & 32.1 \\
R@15 (ceil.) & 80.2 & 37.8 & 78.4 & 70.4 \\
\midrule
\rowcolor{mygray}
\textbf{\ours{}-7B} & \textbf{63.6} & 28.7 & 54.8 & 35.5 \\
\rowcolor{mygray}
\textbf{\ours{}-8B} & 62.0 & \textbf{29.5} & \textbf{55.7} & \textbf{36.8} \\
\bottomrule
\end{tabular}
\end{minipage}
\end{table}

\subsection{Main Results}
\label{sec:main_results}

\subsubsection{4.2.1 Comparison with SOTAs}
We compare \ours{} against three categories of baselines: (1) \textbf{Direct Answer}, where VLMs rely solely on parametric knowledge, to assess the gains from external retrieval; (2) \textbf{Multimodal RAG}, covering diverse retrieval-augmented pipelines, to validate the advantage of visual verification over text-level post-processing; and (3) \textbf{Agentic mRAG}, where models invoke tools through multi-turn reasoning, to demonstrate the necessity of actively verifying retrieval results rather than directly trusting them. We report results for two variants: \ours{}-7B (initialized from Qwen2.5-VL-7B~\cite{bai2025qwen25vltechnicalreport}) and \ours{}-8B (initialized from Qwen3-VL-8B~\cite{bai2025qwen3}). As shown in Tables~\ref{tab:1} and~\ref{tab:mmhops}, both variants consistently surpass all compared methods across all three benchmarks.

\noindent \textbf{Notable improvements on E-VQA.} \ours{}-8B achieves 55.9\% Single-Hop and 54.2\% overall accuracy, outperforming the runners-up QKVQA~\cite{ye2026qkvqa} (+2.2) and ReAG~\cite{compagnoni2025reag} (+7.2), respectively. The E-VQA knowledge base contains ${\sim}$2M Wikipedia pages with Recall@1 of only 15.4\% (Table~\ref{tab:retrieval}), demonstrating that visual reranking becomes critical when the initial retrieval is unreliable.

\noindent \textbf{Consistent superiority on InfoSeek.} \ours{}-7B achieves 50.2\% overall accuracy, surpassing the previous best ReAG~\cite{compagnoni2025reag} (+3.0). Notably, ReAG relies on multi-stage training with SFT cold-start, an external critic module, and RL, whereas \ours{} attains superior performance through RL training alone. Furthermore, MMhops-R1~\cite{zhang2026mmhops}, a concurrent agentic method without visual verification, reaches only 33.2\% on InfoSeek, confirming that visual verification is indispensable for fine-grained entity identification.

\noindent \textbf{Pronounced advantages on multi-hop reasoning.} As shown in Table~\ref{tab:mmhops}, \ours{}-8B achieves 67.2\% on Bridging (+13.2) and 39.8\% on Comparison (+10.4), substantially surpassing the closed-source Gemini-2.5-Pro~\cite{comanici2025gemini} (54.0\% / 29.4\%). These results validate the strong generalization of visual reranking and active rejection to multi-image, multi-hop reasoning.

\begin{figure}[t]
\centering
\includegraphics[width=\textwidth]{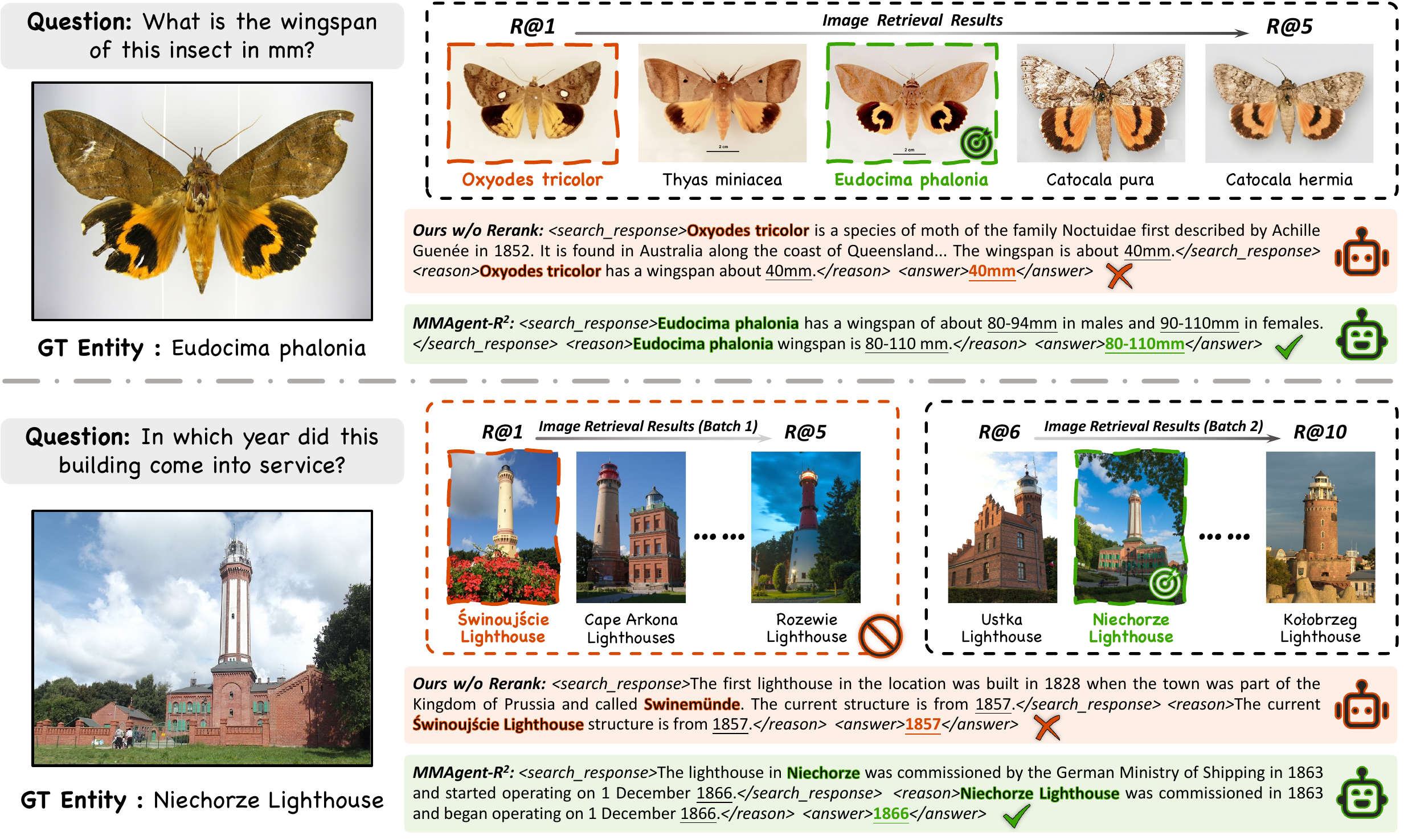}
\caption{Qualitative examples of \ours{}. Top: E-VQA. Bottom: InfoSeek.}
\label{fig:evqa_infoseek}
\end{figure}

\begin{figure}[t]
\centering
\includegraphics[width=\textwidth]{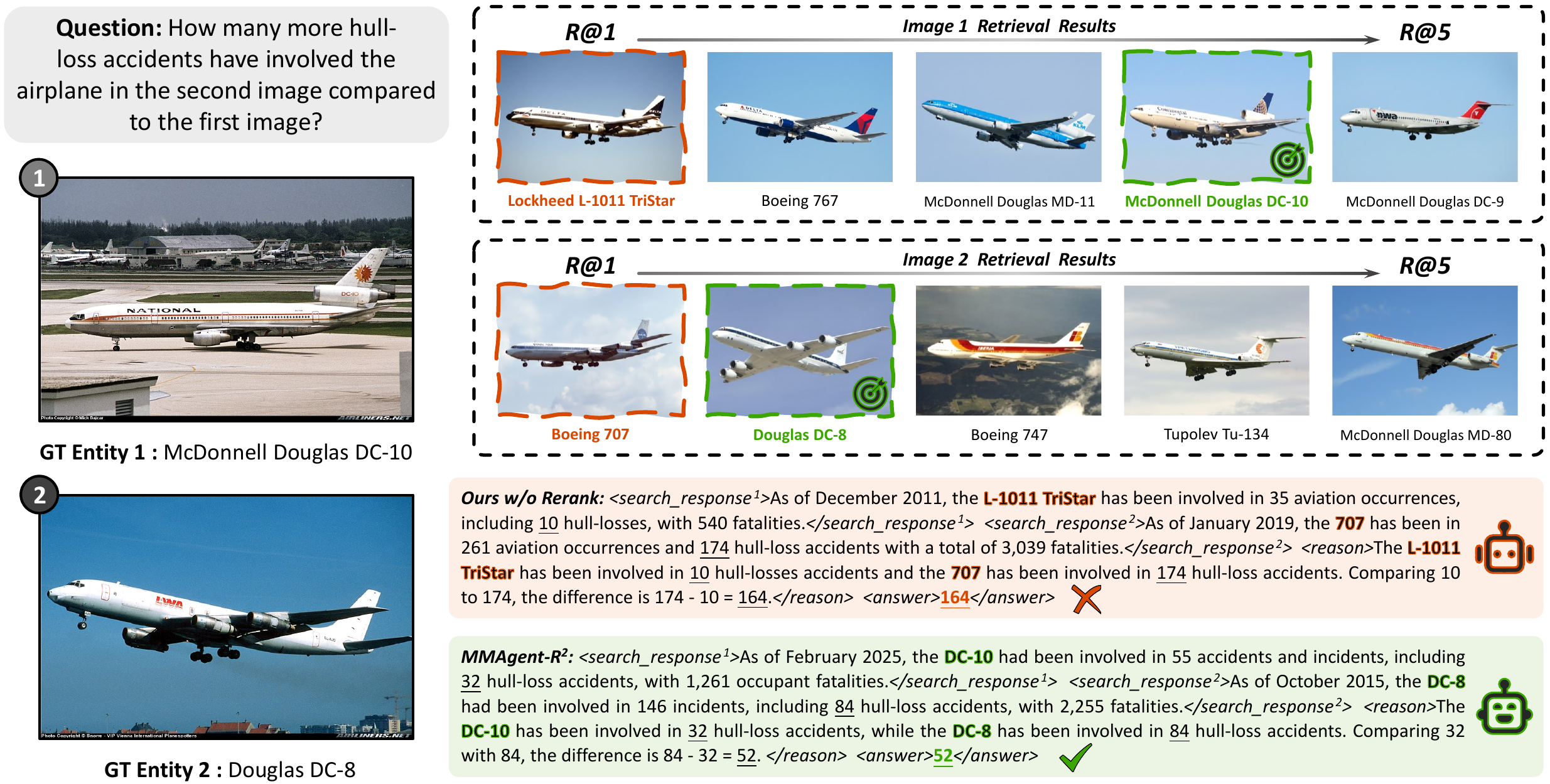}
\caption{Qualitative example of \ours{} on MMhops comparison question.}
\label{fig:mmhops}
\end{figure}

\subsubsection{4.2.2 Improvement on Visual Entity Identification}
\label{sec:rerank_retrieval}
Entity identification accuracy (Ident.) denotes the proportion of samples for which the model correctly identifies the ground-truth entity. For MMhops, which involves multiple query images per sample, a sample is correct only when all query images are simultaneously matched.
As shown in Table~\ref{tab:retrieval}, \ours{} substantially outperforms the visual retriever's Recall@1 across all benchmarks. Notably on E-VQA, where the retriever's Recall@1 is only 15.4\%, \ours{}-8B raises Ident. to 29.5\% (+14.1), indicating that the benefit of visual verification is most pronounced in challenging retrieval scenarios. On InfoSeek, \ours{}-7B raises Ident. from 53.7\% to 63.6\% (+9.9). On MMhops, \ours{}-8B achieves 55.7\% on Bridging (+5.6) and 36.8\% on Comparison (+4.7). These consistent improvements confirm that visual comparison significantly enhances entity identification beyond the retriever's global feature matching. Meanwhile, the remaining gap relative to the upper bound (R@15) suggests that discrimination among a large number of visually similar entities remains challenging.

\subsection{Qualitative Results}
\label{sec:qualitative}
Fig.~\ref{fig:evqa_infoseek} presents qualitative examples on E-VQA~\cite{mensink2023encyclopedic} and InfoSeek~\cite{chen2023can}. In the E-VQA case, the moth candidates share similar coloration and wing shapes, differing only in subtle markings; \ours{} correctly identifies the target entity through visual comparison, whereas the variant without reranking selects the incorrect top-1 candidate, leading to a wrong answer. In the InfoSeek case, the building candidates exhibit similar spire structures, and the correct entity is absent from the first batch; \ours{} actively rejects this batch and successfully identifies the target in the subsequent retrieval. Fig.~\ref{fig:mmhops} illustrates a comparison question from MMhops~\cite{zhang2026mmhops}, where the model must simultaneously identify entities in both query images and retrieve relevant information for reasoning. These cases demonstrate that visual reranking and rejection effectively prevent retrieval errors from propagating to the final answer.

\subsection{Ablation Studies}
\label{sec:ablation}

\begin{table}[t]
\centering
\setlength{\tabcolsep}{3pt}
\renewcommand{\arraystretch}{1.0}

\begin{minipage}[t]{0.52\linewidth}
\centering
\caption{Effect of reranking and rejection. Ident.: entity identification accuracy. U-Q / U-E: Unseen-Question / Unseen-Entity.}
\label{tab:ablation_module}
\scriptsize
\begin{tabular}{lcccc}
\toprule
Variant & Ident. & U-Q & U-E & All \\
\midrule
w/o Rerank, w/o Reject & 53.7 & 40.5 & 39.9 & 40.2 \\
w/o Reject & 57.5 & 44.4 & 44.3 & 44.3 \\
\rowcolor{mygray}
w/ Rerank + Reject & 58.6 & 45.9 & 45.2 & 45.5 \\
\bottomrule
\end{tabular}
\end{minipage}\hfill
\begin{minipage}[t]{0.46\linewidth}
\centering
\caption{Effect of candidate batch size. Avg.~Rej.: average rejection count per sample.}
\label{tab:ablation_batch}
\scriptsize
\begin{tabular}{@{}cccccc@{}}
\toprule
$K$ & Ident. & Avg. Rej. & U-Q & U-E & All \\
\midrule
3 & 58.0 & 0.63 & 45.1 & 44.3 & 44.7 \\
\rowcolor{mygray}
5 & 58.6 & 0.59 & 45.9 & 45.2 & 45.5 \\
7 & 58.7 & 0.51 & 46.0 & 45.4 & 45.7 \\
\bottomrule
\end{tabular}
\end{minipage}
\end{table}

\begin{table}[t]
\centering
\setlength{\tabcolsep}{3pt}
\renewcommand{\arraystretch}{1.0}
\begin{minipage}[t]{0.55\linewidth}
\centering
\caption{Effect of maximum rejection count. }
\label{tab:ablation_reject}
\scriptsize
\begin{tabular}{cccccc}
\toprule
$\text{Rej}_{\max}$ & Ident. & Avg. Rej. & U-Q & U-E & All \\
\midrule
0 & 57.5 & 0 & 44.4 & 44.3 & 44.3 \\
1 & 58.3 & 0.46 & 45.2 & 44.9 & 45.0 \\
\rowcolor{mygray}
2 & 58.6 & 0.59 & 45.9 & 45.2 & 45.5 \\
3 & 58.1 & 0.74 & 45.1 & 44.5 & 44.8 \\
\bottomrule
\end{tabular}
\end{minipage}\hfill
\begin{minipage}[t]{0.43\linewidth}
\centering
\caption{Effect of rejection reward weight $\delta$ in Eq.~\eqref{eq:step_reward}.}
\label{tab:ablation_delta}
\scriptsize
\begin{tabular}{cccccc}
\toprule
$\delta$ & Ident. & Avg. Rej. & U-Q & U-E & All \\
\midrule
0.1 & 57.9 & 0.42 & 45.2 & 44.6 & 44.9 \\
\rowcolor{mygray}
0.2 & 58.6 & 0.59 & 45.9 & 45.2 & 45.5 \\
0.3 & 58.4 & 0.68 & 45.5 & 44.9 & 45.2 \\
\bottomrule
\end{tabular}
\end{minipage}
\end{table}

We train \ours{}-8B ablation variants on 26K samples from InfoSeek to verify the contribution of each key component.

\noindent \textbf{Effect of Rerank and Reject.}
As shown in Table~\ref{tab:ablation_module}, removing both mechanisms reduces the model to answering based on the retriever's top-1 result (Ident.~53.7\%, VQA accuracy~40.2\%). Adding visual reranking improves Ident. and VQA accuracy to 57.5\% and 44.3\% (+3.8/+4.1), confirming visual comparison as the primary driver. Further incorporating rejection reaches 58.6\% and 45.5\%, showing that rejection provides additional gains by expanding candidate coverage. The two mechanisms thus improve entity identification from complementary dimensions: precision and candidate coverage.

\noindent \textbf{Effect of Candidate Batch Size $K$.}
Table~\ref{tab:ablation_batch} shows the effect of the candidate batch size $K$, \ie, the number of candidates presented per reranking step. At $K{=}3$, the limited per-step coverage forces more frequent rejections (Avg.~Rej.~0.63), resulting in lower VQA accuracy (44.7\%). At $K{=}7$, broader coverage leads to fewer rejections (0.51) and marginally improves accuracy (45.7\%), but at higher token cost with diminishing returns. $K{=}5$ strikes the best balance between accuracy and efficiency.

\noindent \textbf{Effect of Maximum Rejection Count.}
As shown in Table~\ref{tab:ablation_reject}, performance improves steadily as the rejection limit increases from 0 to 2 (VQA accuracy from 44.3\% to 45.5\%), but declines at 3 (44.8\%), as redundant candidates lengthen the context and degrade reasoning quality. Notably, average rejections remain well below the allowed maximum (\eg, only 0.59 at $\text{Rej}_{\max}{=}2$), indicating that the model learns to reject on demand rather than expanding indiscriminately.

\noindent \textbf{Effect of Rejection Reward Weight $\delta$.}
As shown in Table~\ref{tab:ablation_delta}, performance peaks at $\delta{=}0.2$ (VQA accuracy 45.5\%): a weaker incentive ($\delta{=}0.1$) under-utilizes the rejection mechanism, yielding lower accuracy (44.9\%), while a stronger one ($\delta{=}0.3$) triggers unnecessary rejections that lengthen the reasoning context and slightly degrade accuracy to 45.2\%. A moderate $\delta$ thus best balances encouraging necessary and suppressing excessive rejections.

\section{Conclusion}
\label{sec:conclusion}
We propose \ours{}, which equips agentic mRAG with visual-level verification over retrieved candidates by introducing visual reranking and active rejection. Combined with a composite reward function incorporating step-level verification rewards and GRPO optimization, \ours{} achieves state-of-the-art performance on multiple KB-VQA benchmarks, with particularly notable gains in challenging retrieval scenarios and multi-image multi-hop reasoning tasks. Our results suggest that empowering models with active verification and error correction capabilities is a key direction for advancing knowledge-based VQA.

\noindent \textbf{Limitations.} The performance of \ours{} is still bounded by the retriever's recall ceiling, and injecting numerous candidate images into the reasoning context incurs additional token overhead. Future work may explore compressing candidate image information while preserving perceptual precision, as well as joint optimization with the retriever.

\section*{Acknowledgements}
This research was supported by multiple funding sources, including the New Generation Artificial Intelligence-National Science and Technology Major Project (2025ZD0123401), the National Natural Science Foundation of China (U24A20331, 62302501, 62192782, 62532015, 62536002, U2441241), Beijing Natural Science Foundation (L251005, L223003, L243015, L252032), and Beijing Major Science and Technology Project (Z251100008425008).

\bibliographystyle{splncs04}
\bibliography{main}

\clearpage
\appendix

\setcounter{section}{0}
\setcounter{figure}{0}
\setcounter{table}{0}
\setcounter{algorithm}{0}
\setcounter{equation}{0}

\renewcommand{\thesection}{\Alph{section}}
\renewcommand{\thesubsection}{\thesection.\arabic{subsection}}
\renewcommand{\thefigure}{S\arabic{figure}}
\renewcommand{\thetable}{S\arabic{table}}
\renewcommand{\thealgorithm}{S\arabic{algorithm}}
\renewcommand{\theequation}{S\arabic{equation}}

\begin{center}
{\Large\bfseries Supplementary Material}\\[0.8em]
{\large\bfseries \ours{}: Learning to Rerank and Reject for Agentic mRAG}
\end{center}
\vspace{1em}

\section{Agentic Reasoning Loop}
\label{sec:supp_agent_loop}
This section presents the complete pseudocode (Algorithm~\ref{alg:agent_loop}) for the multi-turn agentic reasoning loop described in Section~3.2.2. At each turn, the policy $\pi_\theta$ reasons over the accumulated interaction history $\mathcal{H}_{t-1}$ and selects an action from the action space $\mathcal{A}$. The environment executes the action, returns the corresponding observation, and appends it to the history to form a continuously updated evidence chain. Each query image is subject to a maximum rejection count $\text{Rej}_{\max}$, beyond which a boundary prompt is issued. The loop terminates when the agent outputs a final answer or reaches the maximum turn limit $T_{\max}$.

\begin{algorithm}[t]
\caption{Agentic Reasoning Loop of \ours{}}
\label{alg:agent_loop}
\begin{algorithmic}[1]
\REQUIRE Query images $\{I_q^j\}_{j=1}^N$, question $Q$, knowledge bases $\mathcal{KB}_v$/$\mathcal{KB}_t$, policy $\pi_\theta$, batch size $K$, max rejections $\text{Rej}_{\max}$, max turns $T_{\max}$
\ENSURE Answer $y$
\STATE $\mathcal{H}_0 \gets \textsc{SystemPrompt} \oplus \textsc{UserPrompt}(\{I_q^j\}, Q)$
\STATE $\text{RejCount}(j) \gets 0$ for all $j \in \{1, \dots, N\}$
\STATE $j_{\text{cur}} \gets \emptyset$ \hfill $\triangleright$ \textit{Currently active query image index}
\FOR{$t = 1$ \TO $T_{\max}$}
    \STATE $\langle \text{reason}_t, a_t \rangle \gets \pi_\theta(\mathcal{H}_{t-1})$ \hfill $\triangleright$ \textit{Autonomous reasoning \& action selection}
    \STATE $o_t \gets \emptyset$ \hfill $\triangleright$ \textit{Environment observation (default: none)}
    \IF{$a_t = \texttt{Answer}(y)$}
        \RETURN $y$ \hfill $\triangleright$ \textit{Terminate: output final answer}
    \ELSIF{$a_t = \texttt{SearchImage}(j)$}
        \STATE $j_{\text{cur}} \gets j$
        \STATE $o_t \gets \textsc{Retrieve}(I_q^j,\, \mathcal{KB}_v,\, K)$ \hfill $\triangleright$ \textit{Top-$K$ candidates from visual index}
    \ELSIF{$a_t = \texttt{Rerank}(e_k)$}
        \STATE Environment records entity $e_k$ \hfill $\triangleright$ \textit{Confirmed entity for subsequent reasoning}
    \ELSIF{$a_t = \texttt{Reject}()$}
        \IF{$\text{RejCount}(j_{\text{cur}}) < \text{Rej}_{\max}$}
            \STATE $o_t \gets \textsc{Retrieve}(I_q^{j_{\text{cur}}},\, \mathcal{KB}_v,\, K)$ \hfill $\triangleright$ \textit{Next non-overlapping batch}
            \STATE $\text{RejCount}(j_{\text{cur}}) \gets \text{RejCount}(j_{\text{cur}}) + 1$
        \ELSE
            \STATE $o_t \gets \textsc{BoundaryPrompt}$ \hfill $\triangleright$ \textit{Rejection limit reached}
        \ENDIF
    \ELSIF{$a_t = \texttt{SearchText}(\{q_i\}_{i=1}^m)$}
        \STATE $o_t \gets \textsc{Retrieve}(\{q_i\},\, \mathcal{KB}_t)$ \hfill $\triangleright$ \textit{Relevant passages per query}
    \ENDIF
    \STATE $\mathcal{H}_t \gets \mathcal{H}_{t-1} \oplus \langle \text{reason}_t, a_t \rangle \oplus o_t$ \hfill $\triangleright$ \textit{Append to history}
\ENDFOR
\RETURN Forced answer from $\mathcal{H}_{T_{\max}}$ \hfill $\triangleright$ \textit{Timeout}
\end{algorithmic}
\end{algorithm}

\section{Prompt Template}
\label{sec:prompt}
This section presents all prompt templates used during the reasoning process of \ours{}, including the system prompt, task instruction, and the boundary prompt issued when the rejection limit is reached.

\begin{tcolorbox}[
    colback=lightgray,
    colframe=black!60,
    title={\textbf{System Prompt}},
    fonttitle=\small,
    left=4pt, right=4pt, top=4pt, bottom=4pt,
    fontupper=\small\ttfamily
]
You are a helpful and harmless assistant.
\end{tcolorbox}

\begin{tcolorbox}[
    colback=lightgray,
    colframe=black!60,
    title={\textbf{Task Instruction Template}},
    fonttitle=\small,
    breakable,
    left=4pt, right=4pt, top=4pt, bottom=4pt,
    fontupper=\small\ttfamily
]
You will be given one or more query images. Answer the given question.
You must conduct reasoning inside <reason> and </reason> first every time you get new information, and on EVERY turn you must start your reply with a <reason>...</reason> block before producing any other tag or content.
For EACH query image, you should call the image\_search tool with the corresponding image\_index to retrieve candidate images. For example, for the first query image:
\texttt{<search\_call>\{"name": "image\_search", "arguments": \{"image\_index": 1\}\}</search\_call>}
After receiving candidate images, first describe the key visual features of the query image, then compare each candidate against it one by one. If you find a candidate whose entity matches the query image, output \texttt{<rerank>ENTITY\_NAME</rerank>} to select it. If NONE of the candidates in the current batch correspond to the query image, output \texttt{<rerank>Reject</rerank>} to reject the entire batch and retrieve more candidates.
You may also call text\_search to gather additional information:
\texttt{<search\_call>\{"name": "text\_search", "arguments": \{"query\_list": ["your query"]\}\}</search\_call>}
When you are ready to respond, output your final answer inside \texttt{<answer>...</answer>} (a single word or phrase). For example, \texttt{<answer>Beijing</answer>}.
Question: [USER QUESTION]
\end{tcolorbox}

The template enables multi-turn interaction through the following stop strings: \texttt{</rerank>}, \texttt{</answer>}, and \texttt{</search\_call>}. Whenever the model generates a stop string, the inference engine pauses generation, the environment parses and executes the corresponding action, and then appends the feedback to the context before continuing the next generation turn.

When the rejection count for a query image reaches $\text{Rej}_{\max}$, the environment returns the following boundary prompt instead of a new candidate batch:

\begin{tcolorbox}[
    colback=lightgray,
    colframe=black!60,
    title={\textbf{Boundary Prompt (Rejection Limit Reached)}},
    fonttitle=\small,
    left=4pt, right=4pt, top=4pt, bottom=4pt,
    fontupper=\small\ttfamily
]
No more candidate images available. Please select the best matching entity from the current candidates for reranking.
\end{tcolorbox}

\section{Additional Implementation Details}
\label{sec:impl_details}

\subsection{Retrieval System}
\label{sec:retrieval_system}

\noindent \textbf{Image Retrieval.}
We use EVA-CLIP-8B to jointly encode the reference images and titles of each entry in the Wikipedia knowledge base. Specifically, images are encoded by EVA-CLIP-8B's vision encoder and titles by its text encoder; the two feature vectors are concatenated to form a unified representation for each entry. All entry features are indexed using Faiss-GPU with cosine similarity. At retrieval time, the query image is encoded by the same vision encoder, and the top-$K$ most similar entries are returned as a single candidate batch. To support candidate expansion after rejection, we pre-compute available candidates for each query image. During training and inference, these candidates are served in non-overlapping batches of size $K$ (default $K{=}5$). Consecutive retrieval calls triggered by rejection return subsequent batches.

\noindent \textbf{Text Retrieval.}
We build a dense vector index over Wikipedia articles using E5. Wikipedia texts are segmented into 800-character chunks before encoding, and each text query returns the top-3 most relevant chunks. The text retrieval service is implemented as a local FastAPI server, serving both training and inference.

\subsection{Environment Response Masking}
\label{sec:masking}
Expanding on Section~3.3.2 of the main paper, during multi-turn rollout training, all environment response tokens are masked, ensuring that policy gradient updates are applied exclusively to the ``Reasoning'' and ``Action'' tokens generated by the model itself. The masked environment tokens include: (1)~candidate image tokens returned by image retrieval (both visual tokens and entity name text); (2)~passage tokens returned by text retrieval; (3)~confirmation messages and boundary prompts from the environment. Only tokens generated by the model within \texttt{<reason>} blocks and action tags (\texttt{<search\_call>}, \texttt{<rerank>}, \texttt{<answer>}) contribute to the policy gradient computation.

\section{Dataset Details}
\label{sec:dataset_details}

\subsection{Training Data Construction}
\label{sec:data_construction}
For each dataset, we sample a subset from the original training split rather than using the full data. The key preprocessing step is candidate filtering: we retain only samples where the ground-truth entity is reachable within the allowed number of rejections, \ie, the correct entity appears in the candidate pool accessible through the initial retrieval and at most $\text{Rej}_{\max}$ rejection-triggered retrievals. This ensures the effectiveness of verification reward signals during training, as the model has at least the possibility of receiving positive feedback through correct reranking or rejection decisions. Notably, this filtering is \textbf{not} applied during evaluation; all test samples are evaluated, including difficult cases where the ground-truth entity falls outside the reachable candidate pool.

\subsection{Dataset Statistics}
\label{sec:data_stats}
Table~\ref{tab:dataset_stats} summarizes the key statistics of the three benchmark datasets. InfoSeek and MMhops share a knowledge base of 100K Wikipedia pages, while E-VQA employs a much larger 2M-page knowledge base, resulting in significantly higher retrieval difficulty. The Bridging subset of MMhops requires single-image input, while Comparison requires multi-image input; both demand 3--4 reasoning steps.

\begin{table}[t]
\centering
\caption{Dataset statistics and knowledge base configurations.}
\label{tab:dataset_stats}
\small
\begin{tabular}{lccc}
\toprule
\textbf{Statistic} & \textbf{InfoSeek} & \textbf{E-VQA} & \textbf{MMhops} \\
\midrule
Training samples & 65K & 67K & 17K \\
Evaluation samples & 73.7K & 5.7K & 6.2K \\
Knowledge base size & 100K pages & 2M pages & 100K pages \\
Query images per sample & 1 & 1 & 1--2 \\
Reasoning hops & Single & Single, 2-hop & 3--4 hops \\
Evaluation metric & VQA Acc. & BEM score & VQA Acc. \\
\bottomrule
\end{tabular}
\end{table}

\end{document}